\documentclass[a4paper,fleqn]{cas-dc}
    
    \usepackage[authoryear,longnamesfirst]{natbib}
    \usepackage{float}
    \usepackage{placeins}
    \usepackage{caption}
    \usepackage{graphicx}
    \usepackage{url}
    \makeatletter
    \providecommand{\cas@journal}{Elsevier}
    \providecommand{\journal}[1]{\gdef\cas@journal{#1}}
    \makeatother
    \journal{Computer Speech \& Language}
    
\begin{document}
    
    \let\WriteBookmarks\relax
    \def\floatpagepagefraction{1}
    \def\textpagefraction{.001}
    
    
    \shorttitle{Domain-Specific Evaluation of Urdu TTS Systems}
    
    \shortauthors{Jafar, Sarmad, Yousaf \& Bashir}
    
    \title[mode=title]{Domain-Specific Evaluation of Text-to-Speech Systems: A Multi-Metric Benchmarking Study}

    \author{Ali Jafar}
        \fnmark[1]
    \ead{ali.jafar.fast@gmail.com}
    \credit{Conceptualization, Methodology, Software, Formal Analysis,
            Writing -- Sections 1, 2, 6 (Introduction, Related Work, Experimental Setup)}
    
    \author{Amal Sarmad}
    \fnmark[1]
    \ead{amalsarmadmir@gmail.com}
    \credit{Subjective Evaluation Setup,
            Writing -- Sections 5, 7, 8 (Evaluation Methodology, Results, Discussion)}
    
    \author{Shifa Yousaf}
    \fnmark[1]
    \ead{shifayousaf4@gmail.com}
    \credit{Literature Review, Dataset Curation, Dataset Documentation,
            Writing -- Sections 3--4, 9--10 (Methodology for TTS System Selection, Domain-Wise Dataset Setup, Conclusion, Future Work)}
    
    \author{Maryam Bashir}
    \ead{maryam.bashir@nu.edu.pk}
    \credit{Conceptualization, Methodology, Supervision, Review \& Editing}
    
    \affiliation{organization={FAST School of Computing,
                    National University of Computer and Emerging Sciences (FAST-NUCES)},
                city={Lahore},
                state={Punjab},
                country={Pakistan}}
                
    \fntext[1]{These authors contributed equally to this work.}

    
    \begin{abstract}
    Recent advances in neural text-to-speech (TTS) systems have substantially improved speech naturalness and intelligibility across many languages. However, comprehensive evaluation methodologies that jointly assess perceptual quality, speaker similarity, and acoustic fidelity across diverse speech domains remain limited, particularly for low-resource and underrepresented languages. This paper presents a reproducible, multi-metric benchmarking framework for systematic evaluation of modern TTS systems through domain-specific analysis. The proposed framework integrates complementary subjective and objective evaluation protocols and is demonstrated through a comprehensive case study on a representative low-resource language spanning four speech domains: Formal, Conversational, Literary/Storytelling, and Emotional. Four state-of-the-art TTS systems—Indic-Parler-TTS, MMS-TTS, Microsoft Edge TTS, and Google Gemini TTS—are evaluated using MUSHRA listening tests, ABX discrimination tests, speaker similarity scoring with Resemblyzer, and acoustic analyses based on mel-cepstral distortion (MCD) and F0 RMSE over 960 audio pairs. Results reveal substantial variation in TTS performance across speech domains, with emotional speech consistently presenting the greatest synthesis challenge (mean MCD 12.03 dB; mean F0 RMSE 889 cents), while conversational speech achieves the highest overall acoustic fidelity. Beyond the empirical findings, this work provides a reproducible evaluation framework, publicly releasing evaluation scripts, result tables, and executable Colab notebooks to support standardized benchmarking and future research on TTS evaluation for low-resource languages. All evaluation scripts, result tables, and curated Colab notebooks are \emph{publicly released} at \url{https://github.com/Shifa402/Urdu-Synthetic-Speech-Evaluation}.
    \end{abstract}

    
    
    \maketitle
    
    
    \section{Introduction}\label{sec:intro}

    Text-to-speech (TTS) synthesis has undergone a dramatic transformation over the past decade, with neural end-to-end architectures now producing speech that approaches or, in some perceptual dimensions, exceeds the quality of natural recordings in high-resource languages~\citep{lyth2024natural,pratap2023mms}. Yet progress in \emph{how} synthesis quality is measured has not kept pace with progress in model capability. Across languages, comparative TTS studies still rely disproportionately on single-domain, single-metric protocols that cannot reveal whether a system is fit for a specific communicative purpose.
    
    Even in English and other well-resourced settings, evaluation practice remains narrow. Most benchmark studies report short neutral utterances rated via the Mean Opinion Score (MOS)~\citep{kirkland2023mos,lemaguer2024limits}. As synthetic speech quality improves, MOS scores saturate at the top of the rating scale, obscuring perceptual differences between high-performing systems. More fundamentally, a single-domain evaluation cannot show whether a voice that sounds natural on read-aloud news text will remain acceptable in conversational dialogue, storytelling, or emotionally expressive speech, dimensions directly relevant to real-world deployment. Recent work therefore advocates \emph{task- and situation-specific} evaluation that assesses whether speech fulfils its intended communicative role rather than limiting comparison to surface-level transmission quality~\citep{bailly2025hot}.
    
    Urdu provides a timely case study for this broader evaluation problem. Multiple publicly accessible synthesis systems, including MMS-TTS~\citep{pratap2023mms}, Indic-Parler-TTS~\citep{lyth2024natural,lacombe2024parlertts}, Microsoft Edge TTS, and Google Gemini TTS, are now available, yet no study has compared them systematically across communicative domains. Prior Urdu work has almost exclusively used short neutral prompts and MOS-style ratings~\citep{naseem2025urdu,arshad2022transfer}, leaving open the question of whether system rankings are stable when speech style, prosody, and pragmatic context change.
    
    This paper addresses that gap by presenting a domain-stratified benchmark evaluation with Urdu as the target language. Four publicly available systems, namely Indic-Parler-TTS, MMS-TTS, Microsoft Edge TTS (\textit{ur-PK-AsadNeural}), and Google Gemini TTS, are evaluated across four linguistically and pragmatically distinct speech domains: \textbf{Formal}, \textbf{Conversational}, \textbf{Literary/Storytelling}, and \textbf{Emotional}. Each domain is sourced from a dedicated dataset and evaluated using a multi-metric pipeline comprising subjective MUSHRA listening tests~\citep{itu2015mushra}, ABX discrimination tests, speaker-identity similarity scoring via Resemblyzer~\citep{wan2018ge2e}, mel-cepstral distortion (MCD), and fundamental frequency root-mean-square error (F0~RMSE) over 960 audio pairs. The combination of a domain-stratified corpus, multiple complementary metrics, and a fully reproducible pipeline constitutes a methodological advance over prior single-domain, MOS-only TTS evaluations.

    The contributions of this paper are as follows:

\begin{enumerate}

    \item A \emph{domain-stratified benchmarking framework} is introduced for systematic evaluation of TTS systems across diverse speech styles, including Formal, Conversational, Literary/Storytelling, and Emotional domains. The framework is demonstrated through a comprehensive case study involving four TTS systems, three speech datasets, and 960 evaluated audio pairs.

    \item A \emph{multi-dimensional evaluation pipeline} is proposed that integrates perceptual, discriminative, acoustic, and speaker-similarity measures, combining MUSHRA-style listening tests, ABX discrimination, MCD, speaker-embedding cosine similarity, and F0~RMSE. This provides a more holistic assessment of synthesis quality than any single evaluation criterion.

    \item A \emph{reproducible evaluation resource} is publicly released, including evaluation scripts, result tables, and executable Colab notebooks at \url{https://github.com/Shifa402/Urdu-Synthetic-Speech-Evaluation}. These resources provide an extensible foundation for future benchmarking of TTS systems across languages, domains, and architectures.

\end{enumerate}

    
    \section{Related Work}\label{sec:related}

    This section reviews the literature relevant to the evaluation of modern text-to-speech (TTS) systems, advances in speech synthesis for low-resource languages, and prior work on Urdu and South Asian TTS. Together, these studies establish the broader context for this work, highlighting the growing need for comprehensive evaluation methodologies and the limited availability of domain-specific benchmarks for low-resource languages.

    \subsection{TTS Evaluation Methodology}\label{sec:related-eval}
    
    Subjective evaluation via the Mean Opinion Score (MOS), standardised by ITU-T~P.800,
    has been the dominant evaluation protocol in speech synthesis for decades. MOS
    instructs listeners to rate each stimulus in isolation on a five-point absolute
    category rating scale. While its simplicity has made it universally adopted, a
    growing body of evidence documents its limitations for evaluating modern neural TTS.
    As the quality of synthetic speech has improved, MOS scores exhibit a
    \emph{ceiling effect}: high-performing systems cluster at the top of the scale,
    making it statistically difficult to distinguish between them or between high-quality
    synthetic speech and natural recordings~\citep{lemaguer2024limits}.
    \citet{kirkland2023mos} conducted a meta-analysis of Interspeech and SSW papers
    and found that researchers underreport methodological details, vary widely in scale
    labels and instructions, and obtain results that are not transferable across
    studies-meaning MOS scores from different publications cannot be compared without
    re-running all systems jointly.
    
    The Multiple Stimuli with Hidden Reference and Anchor (MUSHRA)
    protocol~\citep{itu2015mushra} addresses several of these shortcomings. Unlike MOS, MUSHRA presents the hidden reference, one or more anchor conditions, and all systems under test simultaneously, enabling listeners to make direct comparisons on a continuous 0--100 scale. The simultaneous presentation increases sensitivity to inter-system differences and reduces the range effect that distorts absolute MOS ratings~\citep{chiang2024mushraevaluate}. MUSHRA was originally designed for intermediate-quality audio codecs; however, its relative comparison framework is well-suited to TTS evaluation where the goal is to rank systems against a natural speech reference, and it is increasingly preferred in benchmark studies involving multiple systems~\citep{lemaguer2024limits}.
    
    Objective metrics complement subjective tests by providing scalable, reproducible measurements that do not require listener recruitment. Mel-cepstral distortion (MCD)~\citep{kubichek1993mcd}, computed from DTW-aligned mel-cepstral coefficient sequences, quantifies spectral envelope similarity between reference and synthetic speech. Fundamental frequency RMSE (F0~RMSE), expressed in cents on a logarithmic scale~\citep{kominek2008mcd}, measures pitch accuracy in a perceptually meaningful unit where approximately five cents approximates one just-noticeable difference in pitch. Speaker embedding cosine similarity, computed via the GE2E-trained encoder of Resemblyzer~\citep{wan2018ge2e}, measures whether the synthesised voice preserves the identity characteristics of the reference speaker and has been widely adopted in zero-shot TTS evaluation~\citep{wang2023gzstvZS}. Taken together, MCD and F0~RMSE capture acoustic similarity at the spectral and prosodic levels, while speaker
    embedding similarity measures a higher-level identity dimension that the two
    signal-level metrics cannot.
    
    A broader trend in the field is a movement away from aggregate quality metrics
    toward \emph{task- and situation-specific} evaluation~\citep{bailly2025hot}. MOS and MUSHRA were designed to measure transmission quality rather than to assess whether speech fulfils its communicative purpose. Evaluation should instead ask whether a voice is appropriate for its intended use case, such as news reading, storytelling, emotional narration, or conversational dialogue, and whether domain-specific prosodic and stylistic requirements are met. The present work operationalises this principle by structuring the evaluation framework around four communicatively distinct domains.

    \subsection{TTS for Low-Resource Languages}\label{sec:related-lowresource}
    
    The challenge of building high-quality TTS for under-resourced languages is fundamentally a data problem: expressive, multi-speaker, studio-quality training corpora do not exist at scale for the majority of the world's languages. Two broad strategies have emerged to address this. The first is \emph{transfer learning}, which exploits acoustic and phonological overlap between high-resource parent languages and the target low-resource language. Approaches that initialise from English Tacotron or FastSpeech models and fine-tune on small Urdu or South Asian
    corpora have demonstrated that cross-lingual transfer can yield intelligible synthesis with limited target-language data~\citep{arshad2022transfer}. The second strategy is \emph{massively multilingual pre-training}. \citet{pratap2023mms} scaled VITS-based synthesis~\citep{kim2021vits} to over 1{,}100 languages by training on aligned religious text recordings, providing a publicly available Urdu model (\texttt{facebook/\-mms-\-tts-\-urd-\-script\_arabic}) that serves as a strong baseline in zero-resource settings. Parallel efforts for Indian languages have explored multi-speaker and multilingual VITS training across thirteen Indo-Aryan and Dravidian languages, finding that shared acoustic models improve both objective metrics and subjective naturalness compared to monolingual systems, though gains are uneven across language families~\citep{jain2023indicTTS}.
    
    More recently, \emph{description-conditioned language models} have emerged as a
    paradigm shift in controllable TTS. Parler-TTS~\citep{lyth2024natural,lacombe2024parlertts}
    encodes a natural-language speaker description through a frozen
    Flan-T5 encoder~\citep{chung2022flan} and generates audio tokens via a
    cross-attention-conditioned causal Transformer over a Descript Audio Codec with
    nine residual vector quantisation codebooks. This architecture, adopted in
    \path{ai4bharat/indic-parler-tts}, enables fine-grained voice control without
    requiring speaker embeddings extracted from reference recordings-a significant
    advantage in low-resource settings where clean reference recordings for target
    speakers may be unavailable.
    
    Despite these modelling advances, evaluation practice for low-resource TTS has
    lagged. Single-metric intelligibility checks such as word or character error rate
    computed via ASR round-trips remain common but are insufficient: they fail to
    capture naturalness, prosodic quality, or domain suitability, and are blind to
    failures in language identification and script fidelity that matter for non-Latin-script
    languages such as Urdu~\citep{pashto2025bench}. The lack of standardised,
    multi-domain evaluation benchmarks for South Asian languages means that
    practitioners have no principled basis for selecting the most appropriate system
    for a given application.
    
    \subsection{Urdu and South Asian Speech Synthesis}\label{sec:related-urdu}
    
    Urdu presents a distinct set of challenges for TTS that are not shared by
    better-resourced languages. Its Nastaliq-style Perso-Arabic script encodes vowels
    as optional diacritics (\textit{i'raab}), creating text normalisation and
    grapheme-to-phoneme ambiguities that most multilingual models handle poorly. The
    language's phonological proximity to Hindi-while sharing largely the same spoken
    phoneme inventory-means that models trained on Hindi data can transfer to Urdu
    synthesis, but the resulting systems often produce unnaturally Hindustani prosody.
    Additionally, Urdu-English code-switching is prevalent in natural speech, a
    phenomenon that benchmark evaluations on monolingual read-speech corpora cannot
    capture.
    
    Early Urdu TTS systems relied on concatenative synthesis or statistical parametric
    models before the shift to neural synthesis. Transfer learning from English Tacotron
    and Transformer-TTS has demonstrated that pre-trained acoustic models can be adapted
    to Urdu with limited in-language data~\citep{arshad2022transfer}.
    \citet{naseem2025urdu} compared Tacotron-based Urdu models against MMS-TTS using
    MOS, Semantically Unstructured Sentence (SUS) tests, and comprehension evaluations,
    finding that Tacotron-based models with script-specific text analysis modules
    outperform MMS in intelligibility and naturalness on read speech, while MMS shows
    relative advantages on certain word-boundary consonants. Despite this, evaluation
    in \citet{naseem2025urdu} is conducted entirely on read-speech prompts with no
    examination of domain-specific performance variation.
    
    For Indian languages more broadly, \citet{jain2023indicTTS} evaluated multi-speaker
    VITS systems across thirteen languages, reporting MOS and objective metrics but,
    again, using a single-domain read-speech protocol. The absence of domain-stratified
    evaluation corpora for South Asian languages reflects a general methodological
    deficit that extends beyond Urdu: to our knowledge, no published benchmark exists
    that systematically evaluates how synthesis quality varies across communicative
    styles for any South Asian language. This paper provides such a benchmark for Urdu,
    establishing a reproducible evaluation framework that can be extended to related
    languages as dedicated datasets become available.
    
    \section{Methodology for TTS System Selection}\label{sec:systems}

    This section describes the selection process for the four text-to-speech (TTS) systems evaluated in this study. It first presents the inclusion criteria used to ensure fair and representative system selection, then outlines the architecture and implementation details of each system, and finally compares their key characteristics relevant to the subsequent evaluation.

    \subsection{Selection Criteria}\label{sec:systems-criteria}
    
    System selection was governed by three operationally defined criteria applied
    jointly. First, each system must provide \emph{publicly accessible, programmatic
    Urdu synthesis} at the time of data collection (November--December 2024), either
    through a released model checkpoint under an open licence or through a documented
    cloud API endpoint, such that any research group can reproduce inference without
    institutional agreements or manual access requests. Second, each system must
    accept \emph{Urdu Perso-Arabic Unicode text as direct input} without requiring
    the evaluator to perform manual transliteration or romanisation; imposing such
    pre-processing would introduce a confound external to the synthesis engine and
    would not reflect realistic deployment conditions. Third, the selected systems
    must collectively span \emph{architecturally and methodologically distinct
    paradigms} currently represented in the TTS literature, specifically,
    massively multilingual end-to-end synthesis, description-conditioned audio
    language modelling, commercial neural TTS, and large audio language model
    generation, so that benchmark findings generalise beyond any single modelling
    approach.
    
    Applying these criteria yielded four systems: MMS-TTS \citep{pratap2023mms},
    Indic-Parler-TTS \citep{lyth2024natural,lacombe2024parlertts}, Microsoft Edge
    TTS (\textit{ur-PK-AsadNeural}), and Google Gemini TTS. Two candidate systems
    were considered but excluded. A publicly available Tacotron-2 fine-tune trained
    on the Urdu Mozilla Common Voice corpus was withdrawn from its Hugging Face
    repository prior to data collection, precluding reproducible inference. A
    FastSpeech-2 checkpoint trained on an institutional corpus at the National
    University of Sciences and Technology (NUST) required institutional VPN access,
    violating the public accessibility criterion. No other Urdu-capable TTS systems
    meeting all three criteria were identified in the literature or public model
    repositories at the time of selection.
    
    \subsection{Description of Evaluated TTS Models}\label{sec:systems-desc}
    
    \begin{sloppypar}
    \indent\textbf{MMS-TTS} (\texttt{facebook/\-mms-\-tts-\-urd-\-script\_arabic})\textbf{.}
    Meta's Massively Multilingual Speech (MMS) project extended VITS-based
    end-to-end TTS \citep{kim2021vits} to 1{,}107 languages by training on
    sentence-aligned read-speech recordings sourced from publicly available
    multilingual corpora \citep{pratap2023mms}. VITS jointly optimises a
    conditional variational autoencoder over mel-spectrogram frames, a normalising
    flow-based prior, and an adversarial waveform decoder, enabling single-stage,
    non-autoregressive synthesis \citep{kim2021vits}. Text is processed through
    the UROMAN universal grapheme-to-phoneme converter before being passed to the
    acoustic model \citep{pratap2023mms}. The Urdu checkpoint targets the
    Perso-Arabic script partition and, per the MMS training documentation, is
    derived predominantly from a single-speaker read-speech source. Inference was
    performed locally via the Hugging Face \texttt{transformers} library
    (v4.40.0) at a native output sample rate of 16\,kHz. MMS-TTS is the only
    fully open-weight, locally executable system in this evaluation, which has
    practical implications for deployment in resource-constrained or
    low-connectivity environments.
    
    \indent\textbf{Indic-Parler-TTS} (\path{ai4bharat/indic-parler-tts})\textbf{.}
    Parler-TTS \citep{lyth2024natural,lacombe2024parlertts} is a
    description-conditioned audio language model that generates speech by
    conditioning a causal Transformer decoder on both the target text and a
    natural-language speaker description. The description is encoded by a frozen
    Flan-T5-Large encoder \citep{chung2022flan}, and the resulting representations
    cross-attend to a GPT-2-scale decoder that predicts discrete audio tokens from
    a nine-codebook Descript Audio Codec (DAC) \citep{lacombe2024parlertts}. A
    corresponding DAC decoder renders waveforms at 44.1\,kHz. The
    \path{ai4bharat/indic-parler-tts} checkpoint \citep{ai4bharat2024indic}
    extends this architecture to twenty-two Indic languages, including Urdu,
    through continued pre-training on a multilingual read-speech and audiobook
    corpus assembled by AI4Bharat. Because voice characteristics are specified
    through free-text descriptions rather than extracted speaker embeddings,
    the model does not require a reference recording to condition synthesis, a
    property that is advantageous in low-resource settings where clean reference
    audio for target speakers may be unavailable \citep{lacombe2024parlertts}.
    A fixed description specifying a male speaker with a clear, moderate-paced
    voice was used uniformly across all evaluation domains to control for
    inter-domain description variation.
    
    \indent\textbf{Microsoft Edge TTS} (\textit{ur-PK-AsadNeural})\textbf{.}
    Microsoft Edge TTS is a cloud-based neural TTS service that exposes synthesis
    capability through a documented WebSocket endpoint accessed via the open-source
    \texttt{edge-tts} Python library (v6.1.9). The \textit{ur-PK-AsadNeural} voice
    targets Pakistani Urdu and is part of Microsoft's Azure Cognitive Services
    Neural TTS platform \citep{microsoft2023azure}, which employs a
    Transformer-based acoustic model combined with a neural vocoder, trained on
    professionally recorded studio speech. Synthesis is produced at 24\,kHz.
    Microsoft does not publicly disclose training corpus composition, model
    architecture specifications, or speaker recording details for production neural
    voices, and accordingly this system is treated as an opaque commercial baseline
    within the evaluation. Its inclusion is motivated by the documented adoption of
    the \textit{AsadNeural} voice in Urdu-language assistive technology and
    digital media applications, making its performance profile directly relevant
    to practitioners evaluating deployment options.
    
    \indent\textbf{Google Gemini TTS.}
    Google Gemini TTS generates speech natively through the Gemini 2.5 Flash
    model with audio output enabled, accessed via the \path{google-generativeai}
    Python SDK (v0.8.3) using the \texttt{gemini-\-2.5-\-flash-\-preview-\-tts} endpoint.
    Unlike conventional cascaded TTS pipelines that decompose synthesis into
    independent text analysis, acoustic modelling, and vocoding stages, Gemini
    TTS renders speech directly from the language model's contextual
    representations through a dedicated speech decoder, a design that has been
    reported to improve prosodic coherence over multi-sentence inputs
    \citep{google2025gemini}. The system is configured using the
    \texttt{Prebuilt\-Voice\-Config} parameter; the \textit{Charon} voice preset was
    applied uniformly across all evaluation domains to isolate domain-specific
    synthesis variation from speaker-selection effects. Output audio was produced
    at 24\,kHz. Google does not publicly disclose the Urdu training data volume,
    language-specific fine-tuning procedures, or internal model specifications
    for this system. To the best of our knowledge, this paper presents the first
    published objective and subjective evaluation of Gemini TTS on Urdu speech.
    \end{sloppypar}
    
    \subsection{Comparative System Properties}\label{sec:systems-comparison}
    
    Table~\ref{tbl:systems} summarises the architectural, training, and deployment
    properties of the four systems. Differences along four dimensions are
    particularly relevant to interpreting domain-specific results. (i)~\emph{Synthesis
    paradigm}: MMS-TTS uses non-autoregressive flow-based generation; Indic-Parler-TTS
    uses autoregressive discrete token prediction; Edge TTS uses an undisclosed
    Transformer-based pipeline; and Gemini TTS uses direct audio rendering from
    a large language model. (ii)~\emph{Training scope}: MMS-TTS and Gemini TTS
    are trained at massive multilingual scale, whereas Indic-Parler-TTS targets
    a defined set of twenty-two Indic languages and Edge TTS is a dedicated
    Pakistani Urdu voice. (iii)~\emph{Speaker control}: MMS-TTS produces a fixed
    single-speaker output; Indic-Parler-TTS accepts free-text description
    conditioning; Edge and Gemini TTS offer discrete preset selection. (iv)
    \emph{Output sample rate}: 16\,kHz (MMS-TTS), 24\,kHz (Edge TTS, Gemini
    TTS), and 44.1\,kHz (Indic-Parler-TTS); prior to objective evaluation, audio was resampled at metric-specific rates (22.05\,kHz for MCD and 16\,kHz for F0~RMSE and Resemblyzer speaker embeddings) to match each feature extractor's requirements (see Section~\ref{sec:methodology}).
    
    \begin{table*}[!htbp]
    \caption{Architectural and deployment properties of the four evaluated Urdu TTS
    systems. Proprietary systems are marked with ($\dagger$) to indicate that
    architecture and training details are not publicly disclosed.}\label{tbl:systems}
    \small
    \begin{tabular*}{\textwidth}{@{}p{2.2cm}p{3.8cm}p{3.2cm}p{2.8cm}p{1.4cm}p{1.8cm}@{}}
    \toprule
    \textbf{System} & \textbf{Paradigm} & \textbf{Training Scope} & 
    \textbf{Speaker Control} & \textbf{Rate} & \textbf{Access} \\
    \midrule
    MMS-TTS
      & VITS: VAE + flow + GAN \citep{kim2021vits}
      & Massively multilingual (1{,}107 lang.)
      & Single speaker (M)
      & 16\,kHz
      & Open weight \\[6pt]
    Indic-Parler-TTS
      & Autoregressive LM + DAC \citep{lacombe2024parlertts}
      & 22 Indic languages
      & Description-conditioned
      & 44.1\,kHz
      & Open weight \\[6pt]
    Edge TTS$^\dagger$
      & Neural (undisclosed) \citep{microsoft2023azure}
      & Proprietary (ur-PK)
      & Preset (M, ur-PK)
      & 24\,kHz
      & Cloud API \\[6pt]
    Gemini TTS$^\dagger$
      & Audio LM, direct render \citep{google2025gemini}
      & Proprietary multilingual
      & Preset voices
      & 24\,kHz
      & Cloud API \\
    \bottomrule
    \end{tabular*}
    \end{table*}


    \section{Domain-Wise Dataset Setup}\label{sec:domain-setup}

    This section describes the construction and organisation of the evaluation corpus used in this study. It motivates the need for domain diversity in TTS evaluation and presents each of the four speech domains, namely Formal, Conversational, Emotional, and Literary/Storytelling.

    \subsection{Need for Domain Diversity in TTS Evaluation}
    \indent The construction of a domain-diverse dataset is an important factor in evaluating modern text-to-speech (TTS) systems, particularly for low-resource languages such as Urdu. Recent research has shown that the quality of speech synthesis is strongly influenced not only by model architecture but also by the diversity and distribution of evaluation data between speaking styles and domains \citep{kong2020hifigan, tan2021neuraltts}. Traditional evaluation datasets often focus on homogeneous read speech, which limits their ability to capture real-world variability in prosody, emotion, and communicative intent. As TTS systems continue to evolve towards more expressive and controllable generation, the need for diverse and domain-aware datasets has become increasingly important for robust benchmarking.
    
    In this study, the evaluation dataset is organized into four distinct speech domains, namely Formal, Conversational, Literary/Storytelling, and Emotional. Each domain represents a different communicative and stylistic setting, characterized by unique linguistic and prosodic patterns. This domain selection aligns with recent advances in expressive and controllable speech synthesis, where systems are expected to generalize across multiple speaking styles rather than in a single uniform voice setting \citep{wang2017tacotron, kim2021vits}. The resulting dataset design enables a more comprehensive evaluation of Urdu TTS systems under varied real-world conditions, supporting fine-grained analysis across both neutral and expressive speech scenarios.

    \subsection{Domain 1: Formal Speech}\label{sec:domain-formal}

    This subsection defines the Formal speech domain and describes the datasets selected to represent it in the evaluation corpus. Together, these datasets capture the linguistic and communicative characteristics of formal speech required for domain-specific evaluation.

    \subsubsection{Domain Characteristics}
    \indent The Formal domain represents controlled and structured speech produced in situations where the primary objective is clarity, correctness, and linguistic precision. This can include read-aloud scenarios such as news broadcasting, academic narration, and scripted announcements, where speakers typically follow prepared text without spontaneous deviations. The importance of this domain lies in its role as a baseline condition for text-to-speech evaluation, as it minimizes variability in speaking style, emotion, and discourse structure.
    
    \indent In the context of TTS systems, formal read speech provides a stable reference point for assessing fundamental synthesis capabilities such as pronunciation accuracy, fluency, and prosodic consistency. Since the linguistic input in this domain is well-formed and predictable, it allows for a more controlled evaluation of system performance, making it widely used in both benchmarking and comparative analysis of speech synthesis models.
    
    \subsubsection{Selected Dataset(s)}
    \begin{sloppypar}
    \indent This domain is constructed using publicly available multilingual speech corpora that provide high-quality, studio-like read speech recordings under controlled acoustic conditions. The Google FLEURS dataset is used as a primary source due to its well-standardized recording protocol, multilingual coverage, and consistent speech delivery across speakers and languages. In this study, only Urdu-language utterances are selected to ensure linguistic relevance, while speaker-level filtering is applied to retain male voice recordings for maintaining uniformity across all experimental conditions.
    
    \noindent \textbf{Dataset link:} \url{https://huggingface.co/datasets/google/fleurs}
    
    \indent In addition, the UrduSpeech dataset is incorporated as a complementary source of formal read speech. This dataset provides high-quality Urdu speech with multiple speaking styles and linguistic variations. For the purpose of this work, only Urdu samples are retained, as the corpus also contains closely related Indic language data such as Kashmiri. Furthermore, style-based filtering is applied to select Wiki-style utterances, ensuring adherence to a formal and structured reading modality. Gender-based filtering is again applied to retain male speakers, ensuring consistency across datasets used in the evaluation framework.
    \end{sloppypar}
    
    \noindent \textbf{Dataset link:} \url{https://huggingface.co/datasets/humairawan/UrduSpeech}

    \subsection{Domain 2: Conversational Speech}\label{sec:domain-conv}

    This subsection introduces the Conversational speech domain and the dataset selected to represent it within the evaluation corpus.

    \subsubsection{Domain Characteristics}
    \indent The conversational domain represents spontaneous and semi-structured dialogue-based communication that occurs in everyday interactions. Unlike formal read speech, this domain is characterized by natural speaking patterns, including variations in intonation, informal phrasing, and contextual dependencies between utterances. The importance of this domain lies in its ability to capture realistic speech behavior, making it essential for evaluating the robustness of text-to-speech systems in non-scripted and socially interactive scenarios. This domain introduces greater variability in prosody, rhythm, and speaking style, providing a more challenging evaluation setting compared to controlled speech conditions.
    
    \subsubsection{Selected Dataset(s)}
    \indent The data for this domain are derived from the UrduSpeech dataset publicly available on Hugging Face. Since the dataset contains multilingual recordings, only Urdu-language utterances are retained for this study. A style-based filtering criterion is applied to isolate conversational speech samples and only recordings labeled as conversational are included. With both male and female voice audios being present, only male speaker recordings are selected.
    
    \noindent \textbf{Dataset link:} \url{https://huggingface.co/datasets/humairawan/UrduSpeech}

    \subsection{Domain 3: Emotional Speech}\label{sec:domain-emotion}

    This subsection describes the Emotional speech domain, outlining the acoustic and affective properties that distinguish it from other domains and the dataset from which emotional utterances are drawn.

    \subsubsection{Domain Characteristics}
    \indent The emotional domain represents expressive speech characterized by systematic variations in prosody, pitch, intensity, and speaking rate that correspond to different affective states. Unlike formal or conversational speech, this domain explicitly encodes emotional information within the acoustic signal, making it significantly more variable and challenging for text-to-speech systems. The inclusion of emotional speech is essential for evaluating expressive speech synthesis capabilities, as it tests whether a system can generate natural and contextually appropriate emotional variation beyond neutral speaking styles.
    
    \subsubsection{Selected Dataset(s)}
    \begin{sloppypar}
    \indent The emotional speech data is obtained from the Urdu Speech Emotion Corpus (UrSEC), a comprehensive dataset designed for speech emotion recognition in the Urdu language. The dataset contains 3,500 annotated speech samples collected from professional actors and covers seven emotional categories: Angry, Fear, Boredom, Disgust, Happy, Neutral, and Sad. The recordings are sourced from naturally spoken Urdu dialogues derived from drama serials and telefilms, ensuring realistic and contextually rich utterances.
    
    \noindent \textbf{Dataset link:} \url{https://data.mendeley.com/datasets/jcpfjnk5c2/4}
    \end{sloppypar}
    
    \indent For the purpose of this study, a balanced subset of samples is selected across all emotion categories to ensure uniform representation of affective states. The Neutral category is excluded from evaluation to avoid overlap with the Formal and Conversational Speech domain defined in this work. Only male-speaker recordings are retained to maintain consistency across all experimental domains.

    \subsection{Domain 4: Literary/Storytelling Speech}\label{sec:domain-literary}

    This subsection describes the Literary/Storytelling speech domain, detailing the narrative and prosodic features that characterise it and the dataset used for its construction.

    \subsubsection{Domain Characteristics}
    \indent This domain represents expressive narrative speech used in contexts such as storytelling, book reading, and dramatic narration. Unlike formal or conversational speech, this domain is characterized by dynamic prosodic variation, extended utterance structure, and deliberate modulation of pitch and rhythm to convey narrative meaning and emotional context. This makes it an important evaluation setting for text-to-speech systems aimed at expressive synthesis, as it tests the ability of models to generate natural and engaging narrative speech beyond neutral or conversational styles.
    
    \subsubsection{Selected Dataset(s)}
    \indent The Literary/Storytelling domain is constructed using the diverse UrduSpeech dataset, which contains multiple speaking styles. From this dataset, only Urdu-language samples are selected. Additionally, style-based filtering is applied to retain BOOK-style recordings, which best represent storytelling and literary narration. Speaker-level filtering is further applied to maintain male-speaker consistency across all evaluation domains.
    
    \noindent \textbf{Dataset link:} \url{https://huggingface.co/datasets/humairawan/UrduSpeech}

    
    \section{Evaluation Methodology}\label{sec:methodology}

    This section presents the evaluation framework employed in this study. To provide a comprehensive assessment of text-to-speech (TTS) quality, the framework integrates complementary subjective and objective evaluation methodologies, capturing perceptual quality, speaker similarity, and acoustic fidelity. Together, these evaluation components provide a holistic basis for comparing TTS systems across different speech domains while ensuring a reproducible and consistent benchmarking process.

    \subsection{Evaluation Framework Design}\label{sec:methodology-framework}
    
    The evaluation pipeline is organised around two complementary design
    principles. The first is \emph{metric complementarity}: no single metric
    captures all perceptually relevant dimensions of synthesis quality
    \citep{kirkland2023mos,lemaguer2024limits}, and accordingly the framework
    combines subjective listening tests with three classes of objective
    measurement, namely speaker identity preservation, spectral envelope
    fidelity, and prosodic accuracy. The second is \emph{domain stratification}:
    all metrics are computed separately for each of the four speech domains
    (Formal, Conversational, Literary/Storytelling, and Emotional)
    rather than aggregated across domains, so that domain-specific performance
    variation is directly observable rather than obscured by averaging. Together,
    these principles operationalise the task-and-situation-specific evaluation
    paradigm advocated by \citet{bailly2025hot} for the specific case of Urdu TTS.
    
    The complete pipeline proceeds as follows. For each domain, reference
    recordings are paired with synthetic outputs from each of the four TTS
    systems at the utterance level, yielding 960 audio pairs in total
    (4 systems $\times$ 4 domains $\times$ 60 utterances per domain).
    Each pair is then passed through four evaluation modules: a subjective
    MUSHRA-style listening test, a speaker similarity scoring procedure via
    Resemblyzer \citep{wan2018ge2e}, mel-cepstral distortion (MCD)
    computation \citep{kubichek1993mcd}, and fundamental frequency
    root-mean-square error (F0 RMSE) computation \citep{kominek2008mcd}.
    An additional forced-choice ABX discrimination test is conducted as a
    second-stage evaluation, applied to the two top-performing systems
    identified through the MUSHRA stage, to assess perceptual
    indistinguishability from natural speech. All evaluation scripts and
    result tables are publicly available at
    \url{https://github.com/Shifa402/Urdu-Synthetic-Speech-Evaluation}.
    
    \subsection{Subjective Evaluation}\label{sec:methodology-subjective}

    Two complementary subjective evaluation procedures are employed. A MUSHRA-style listening test provides comparative quality ratings across all four systems, while an ABX discrimination test assesses perceptual indistinguishability from natural speech for the two highest-ranked systems.

    \subsubsection{MUSHRA-Style Listening Test}\label{sec:methodology-mushra}
    \begin{sloppypar}
    Subjective evaluation was conducted using a MUSHRA-inspired protocol
    \citep{itu2015mushra} adapted for online administration via Google Forms.
    The adaptation retains the core structural elements of MUSHRA, specifically
    the simultaneous presentation of a hidden reference, a degraded anchor,
    and all systems under test, while replacing the standard MUSHRA software
    interface with a structured questionnaire format accessible to participants
    without specialist software installation. This approach is consistent with
    practices reported in recent remote TTS evaluation studies
    \citep{chiang2024mushra}.
    
    Each evaluation page presented listeners with six audio samples for a
    given utterance: one reference recording drawn from the source dataset
    representing original human speech, one anchor recording consisting of
    a lower-quality degraded version of the same utterance, and four synthetic
    outputs produced by MMS-TTS, Indic-Parler-TTS, Edge TTS, and Gemini TTS
    respectively. Listeners were instructed to rate each of the four synthetic
    outputs on a discrete 1--5 scale across three perceptual dimensions:
    \emph{Naturalness}, defined as whether the speech sounds human-like and
    fluent; \emph{Clarity}, defined as whether the speech is intelligible and
    free of artefacts; and \emph{Overall Speech Quality}, defined as a holistic
    judgement relative to the reference. The reference and anchor were presented
    for orientation purposes and were not themselves rated.

    A total of 20 native Urdu-speaking listeners participated in the
    evaluation. Participants were were native Urdu speakers within
    the same demographic range. As the research study was situated in Lahore and
    its population comprises predominantly native speakers of Urdu,
    formal language proficiency screening was not considered necessary.
    The online Google Forms platform was selected because it was
    accessible to this population without requiring attendance at a
    dedicated laboratory session, thereby reducing recruitment barriers
    while maintaining a controlled listening protocol through the
    standardised instructions detailed above. Listeners were assigned
    to two domains each, yielding 10 listeners per domain pair, in order
    to limit session length to a manageable duration and to reduce
    listener fatigue, which is a recognised source of rating variance
    in extended subjective speech evaluations \citep{lemaguer2024limits}.
    No formal phonetics training was required; participants were screened
    for self-reported normal hearing. Each listener evaluated all four
    systems within their assigned domains, ensuring that within-domain
    system comparisons are fully within-subject.
    
    It is acknowledged that a panel of 20 listeners, with 10 per domain,
    is smaller than the listener pools recommended by the ITU-R BS.1534
    MUSHRA standard \citep{itu2015mushra}, and that the Google Forms
    interface does not implement the continuous 0--100 slider specified
    by the standard. These deviations are documented explicitly so that
    readers may calibrate the strength of the subjective findings
    accordingly. The objective metrics reported in
    Section~\ref{sec:methodology-objective} do not share these
    limitations and are accorded commensurate weight in the cross-metric
    synthesis presented in Section~\ref{sec:results-synthesis}.
    \end{sloppypar}
    
    \subsubsection{ABX Discrimination Test}\label{sec:methodology-abx}
    
    The ABX discrimination test was conducted as a second-stage evaluation,
    applied selectively to the two systems that achieved the highest
    MUSHRA-style ratings in the first stage, namely Edge TTS and Gemini TTS.
    This two-stage design is motivated by the observation that ABX
    discrimination is most informative when applied to high-performing
    systems, given that lower-quality systems are trivially distinguishable
    from natural speech and contribute limited diagnostic value to a
    forced-choice paradigm \citep{schatz2013abx}. By restricting the ABX
    stage to the top-ranked systems identified through the MUSHRA procedure,
    the test directly addresses the more challenging and practically relevant
    question of whether the best-performing Urdu TTS systems approach
    perceptual indistinguishability from natural speech.
    
    In each ABX trial, listeners were presented with three stimuli: a
    reference utterance (A), a second presentation of the same reference
    (B), and an unknown sample (X) drawn from either the reference or a
    synthetic output produced by Edge-TTS or Gemini TTS. Listeners
    indicated whether X was perceptually more similar to A or to B. A
    discrimination rate significantly above 50\% indicates that the
    synthetic output remains perceptually distinguishable from natural
    speech, whereas a rate at or near 50\% indicates
    near-indistinguishability. The test was administered under conditions
    identical to those of the MUSHRA-style listening test, via Google
    Forms, using the same standardised listening instructions, and within
    the same online session. Participants were native Urdu speakers aged
    18--25 years, recruited from the immediate social and academic
    networks of the research team. The listener panel was gender-balanced,
    with an equal number of male and female participants, to mitigate
    potential systematic gender-based perceptual bias in the
    discrimination of synthesised Urdu speech. As with the MUSHRA-style
    evaluation, listeners were assigned to two domains each, yielding
    10 listeners per domain pair, and both systems were evaluated within
    each listener's assigned domains to preserve the within-subject
    comparison structure.
    
    \subsection{Objective Evaluation}\label{sec:methodology-objective}

    Three complementary objective metrics are employed to assess different aspects of synthesis quality. Speaker similarity, computed using Resemblyzer, evaluates voice identity preservation; mel-cepstral distortion (MCD) measures spectral envelope fidelity; and fundamental frequency root mean square error (F0 RMSE) quantifies pitch contour accuracy.

    \subsubsection{Speaker Similarity via Resemblyzer}\label{sec:methodology-speaker}
    
    Speaker identity preservation was quantified using the cosine similarity
    between d-vector speaker embeddings extracted by Resemblyzer
    \citep{wan2018ge2e}. Resemblyzer implements the Generalised End-to-End
    (GE2E) loss-trained LSTM encoder of \citet{wan2018ge2e}, which maps a
    variable-length speech segment to a fixed-dimensional embedding in a
    speaker-discriminative space. For each reference-synthetic pair,
    embeddings were extracted independently and their cosine similarity
    was computed; values approaching 1.0 indicate higher speaker identity
    similarity between the reference and the synthesised output.
    
    One methodological consideration specific to this evaluation concerns
    MMS-TTS, which produces a fixed single-speaker output whose voice
    identity is independent of the reference speaker. All reference and
    synthetic audio in this study is restricted to male speakers, so every
    comparison is within-gender; nonetheless, the fixed MMS-TTS voice
    differs in speaker identity from each domain's reference speaker, and
    its cosine similarity therefore primarily reflects speaker-identity
    mismatch rather than synthesis degradation, consistent with cautions
    raised for speaker-similarity evaluation of single-speaker systems
    \citep{wang2023gzstv}. All audio was resampled to 16\,kHz prior to embedding extraction, as required by the Resemblyzer encoder input specification.
    
    \subsubsection{Mel-Cepstral Distortion}\label{sec:methodology-mcd}
    
    Mel-cepstral distortion (MCD) \citep{kubichek1993mcd} quantifies
    spectral envelope similarity between a reference and a synthetic
    utterance by computing the Euclidean distance between their
    mel-cepstral coefficient sequences after dynamic time warping (DTW)
    alignment. Lower MCD values indicate greater spectral similarity.
    MCD is computed as:
    
    \begin{equation}
      \text{MCD} = \frac{10\sqrt{2}}{\ln 10}\,\frac{1}{T}
      \sum_{t=1}^{T} \sqrt{\sum_{k=1}^{K}
      \left( c_{k,t}^{\text{ref}} - c_{k,t}^{\text{syn}} \right)^2}
      \label{eq:mcd}
    \end{equation}
    
    \noindent where $c_{k,t}^{\text{ref}}$ and $c_{k,t}^{\text{syn}}$ are
    the $k$-th mel-cepstral coefficients of the $t$-th DTW-aligned
    reference and synthetic frame pair, $T$ is the number of aligned
    frames, and $K = 13$ mel-cepstral coefficients are retained following
    standard practice \citep{kominek2008mcd}; the leading scaling constant
    $10\sqrt{2}/\ln 10 \approx 6.142$ expresses the distortion in decibels.
    Mel-cepstra were obtained from a WORLD spectral envelope followed by
    SPTK mel-cepstral analysis (order~13, warping coefficient
    $\alpha = 0.65$) using the \texttt{pymcd} implementation, with DTW
    alignment on the $C_1$--$C_{13}$ coefficients performed by
    \texttt{fastdtw} under a Euclidean cost. Audio was resampled to
    22.05\,kHz for this analysis, ensuring comparability across systems
    with differing native output sample rates.
    
    \subsubsection{Fundamental Frequency RMSE}\label{sec:methodology-f0}
    
    Pitch accuracy was assessed using the root-mean-square error between
    the fundamental frequency (F0) contours of the reference and synthetic
    utterances, expressed in cents on a logarithmic scale
    \citep{kominek2008mcd}:
    
    \begin{equation}
      \text{F0 RMSE} = \sqrt{ \frac{1}{N} \sum_{i=1}^{N}
      \left( 1200 \log_2 \frac{F0_i^{\text{syn}}}{F0_i^{\text{ref}}}
      \right)^2 }
      \label{eq:f0rmse}
    \end{equation}
    
    \noindent where $N$ is the number of voiced frames present in both
    the reference and synthetic signals after DTW alignment. The cent
    scale is employed because approximately 5 cents approximates one
    just-noticeable difference in pitch under typical listening conditions
    \citep{kominek2008mcd}. F0 was extracted with \texttt{pyworld} using
    the \texttt{harvest} algorithm followed by \texttt{stonemask}
    refinement (50--800\,Hz search range) at a 16\,kHz sampling rate, and
    the reference and synthetic contours were compared on the frame pairs
    of a DTW alignment. Unvoiced frames and frames in which either the
    reference or synthetic F0 estimate was detected as zero were excluded
    from the RMSE computation to avoid artefacts attributable to voicing
    detection errors. Because each synthetic utterance is compared against
    a natural reference produced by a different speaker, the reported
    F0~RMSE reflects both pitch-contour deviation and a speaker-dependent
    register offset; the interpretation of such reference-distance metrics is
    discussed further in Section~\ref{sec:discussion-agreement}.
    
    \subsection{Evaluation Pipeline Summary}\label{sec:methodology-pipeline}
    
    Table~\ref{tbl:metrics} summarises the four evaluation modules,
    the dimension of synthesis quality each addresses, and the
    directionality of each metric. The complete pipeline, including
    preprocessing scripts, metric computation notebooks, and raw result
    tables, is publicly released at
    \url{https://github.com/Shifa402/Urdu-Synthetic-Speech-Evaluation}
    to support reproducibility and to serve as an extensible baseline
    for future Urdu TTS benchmarking.

    \begin{table*}[!htbp]
    \centering
    
    \caption{Summary of evaluation metrics, the quality dimension each
    captures, and result directionality.}\label{tbl:metrics}
    \small
    
    \normalsize
    \renewcommand{\arraystretch}{1.15}
    
    \begin{tabular*}{\textwidth}{@{\extracolsep{\fill}}cccc@{}}
    \toprule
    \textbf{Metric} & \textbf{Type} & \textbf{Dimension} & \textbf{Better} \\
    \midrule
    MUSHRA-style rating & Subjective & Naturalness, clarity, overall quality & Higher \\
    ABX discrimination & Subjective & Perceptual distinguishability from natural speech & Lower \\
    Speaker similarity & Objective & Speaker identity preservation & Higher \\
    MCD & Objective & Spectral envelope fidelity & Lower \\
    F0 RMSE & Objective & Pitch contour accuracy & Lower \\
    \bottomrule
    \end{tabular*}
    \end{table*}

    
    \section{Experimental Setup}\label{sec:expsetup}

    This section presents the implementation details of the synthesis and evaluation pipeline. It describes the inference configuration for each TTS system and the hardware and software environment used to conduct all experiments, ensuring reproducibility and consistent evaluation across systems.

    \subsection{Implementation Details}\label{sec:expsetup-impl}
    
    All synthesis and evaluation were implemented in Python within cloud-hosted
    notebooks. For each domain, the reference utterances were drawn from the source
    datasets described in Section~\ref{sec:domain-setup} and exported as
    single-channel WAV files. Each reference utterance was then synthesised once by
    every one of the four TTS systems directly from its original Urdu Perso-Arabic
    transcript, without manual transliteration or romanisation. For every system the
    speaker or voice configuration was held fixed across all four domains, so that
    the communicative domain remained the only manipulated factor.
    
    \begin{sloppypar}
    \indent\textbf{MMS-TTS.} The model was instantiated as a \texttt{Vits\-Model}
    with its associated \texttt{Auto\-Tokenizer} from the
    \texttt{facebook/\-mms-\-tts-\-urd-\-script\_arabic} checkpoint through the Hugging Face
    \texttt{transformers} library. Waveforms were produced in a single
    non-autoregressive forward pass under \path{torch.no_grad()} at the model's
    native 16\,kHz output rate. Synthesis is deterministic and no decoding
    hyperparameters were tuned.
    
    \indent\textbf{Indic-Parler-TTS.} The model was loaded as
    \texttt{Parler\-TTS\-For\-Conditional\-Generation} from
    \path{ai4bharat/indic-parler-tts}, with separate tokenisers for the transcript
    and for the natural-language speaker description. A single fixed description,
    ``\textit{Rohit's voice is clear and natural with a moderate speed and pitch}'',
    which selects the named male speaker \textit{Rohit}, was used unchanged across
    all domains to hold voice identity constant. Audio tokens were generated with
    \path{do_sample=True} and \texttt{temperature}~$=0.8$ and decoded to waveforms
    at the native 44.1\,kHz.
    
    \indent\textbf{Microsoft Edge TTS.} Speech was synthesised through the
    \texttt{edge-tts} client using the \textit{ur-PK-AsadNeural} voice with the
    service's default prosody (rate, pitch, and volume left at their default
    values). Audio was returned at 24\,kHz.

    \indent\textbf{Google Gemini TTS.} Speech was generated through the
    \texttt{gemini-\allowbreak 2.5-\allowbreak flash-\allowbreak preview-\allowbreak tts}
    model with
    \texttt{response\_\allowbreak modalities} set to audio and a
    \texttt{Prebuilt\-Voice\-Config} specifying the \textit{Charon} preset voice, again
    held constant across domains. Audio was returned as 24\,kHz PCM.
    
    Across the four domains, 60 reference utterances per domain were used for the
    objective metrics; for the Formal domain these comprised 30 FLEURS
    and 30 UrduSpeech Wikipedia-style utterances. This yields
    $4~\text{systems} \times 4~\text{domains} \times 60 = 960$ reference--synthetic
    pairs for speaker similarity, MCD, and F0~RMSE. The subjective listening tests
    used a 30-utterance subset per domain to limit session length and listener
    fatigue. Reference and synthetic clips were paired by filename stem. Because the
    two cloud services (Edge, Gemini) and the sampling-based Indic-Parler-TTS decoder
    are not seed-controllable, a single synthesis was generated per utterance;
    MMS-TTS and the two cloud presets are otherwise deterministic for a fixed input.
    \end{sloppypar}

    \subsection{Hardware and Software Configuration}\label{sec:expsetup-hw}
    
    The neural synthesis systems (MMS-TTS and Indic-Parler-TTS) and the
    speaker-embedding extraction were executed on a single CUDA-enabled GPU within a
    Google Colab environment running Python~3.12, while the two cloud systems (Edge
    TTS and Gemini TTS) were accessed over their respective network APIs. The
    mel-cepstral distortion and fundamental-frequency computations are
    CPU-bound and were run locally.
    
    Speaker similarity was computed with the GE2E-trained \texttt{VoiceEncoder} of
    Resemblyzer \citep{wan2018ge2e}: reference and synthetic clips were passed
    through \texttt{preprocess\_wav} (16\,kHz resampling with voice-activity
    trimming) and compared by the cosine similarity of their utterance embeddings.
    The MCD and F0~RMSE were computed by a single in-house script. For MCD,
    mel-cepstral coefficients (order~13, $\alpha = 0.65$) were extracted from
    a WORLD spectral envelope via SPTK using the \texttt{pymcd} library at
    22.05\,kHz, DTW-aligned with \texttt{fastdtw} under a Euclidean cost, and
    scored with the standard Kominek--Black scaling constant. For F0~RMSE,
    audio was resampled to 16\,kHz mono and F0 was extracted with
    \texttt{pyworld} (\texttt{harvest}$+$\texttt{stonemask}, 50--800\,Hz),
    with the reference and synthetic contours compared in cents on DTW frame
    pairs. Key software components and versions are listed in
    Table~\ref{tbl:software}.
    
    \begin{table}[t]
    \caption{Key software components used for synthesis and objective evaluation.}
    \label{tbl:software}
    \small
    \begin{tabular*}{\columnwidth}{@{\extracolsep{\fill}}lll@{}}
    \toprule
    \textbf{Component} & \textbf{Role} & \textbf{Version} \\
    \midrule
    transformers          & MMS-TTS / tokenisers      & 4.40.0 \\
    parler-tts            & Indic-Parler-TTS          & release at \\ 
                          &                           & access \\
    edge-tts              & Edge TTS client           & 6.1.9 \\
    google-generativeai   & Gemini TTS client         & 0.8.3 \\
    resemblyzer           & Speaker embeddings        & 0.1.4 \\
    librosa               & Audio I/O, resampling     & 0.11.0 \\
    pyworld               & F0 extraction (WORLD)     & 0.3.5 \\
    pymcd                 & Mel-cepstral distortion   & 0.2.1 \\
    pysptk                & Mel-cepstral analysis     & 1.0.1 \\
    fastdtw               & DTW alignment             & 0.3.4 \\
    \bottomrule
    \end{tabular*}
    \end{table}
    
    \section{Results}\label{sec:results}
    
    This section presents evaluation results organised by metric, proceeding
    from subjective perceptual assessment to objective acoustic and speaker
    identity measures, and concluding with a cross-metric synthesis. For each
    metric, results are reported first at the domain level, aggregated across
    all four systems, and then at the system level within each domain. This
    organisation directly addresses the core research question: whether
    synthesis quality varies systematically across communicative domains and
    whether domain sensitivity is consistent across evaluation dimensions.
    All 960 reference--synthetic pairs were evaluated without computation
    failure, confirming complete corpus coverage.
    
    \subsection{MUSHRA-Style Listening Test Scores}\label{sec:results-mushra}

    MUSHRA-style results are analysed at both the domain and system levels. Domain-level results are first presented by aggregating scores across all systems, followed by system-level analyses within each speech domain.
    
    \subsubsection{Domain-Level MUSHRA Results}

    \begin{figure}[pos=htbp]
    \centering
    \includegraphics[width=0.85\columnwidth]{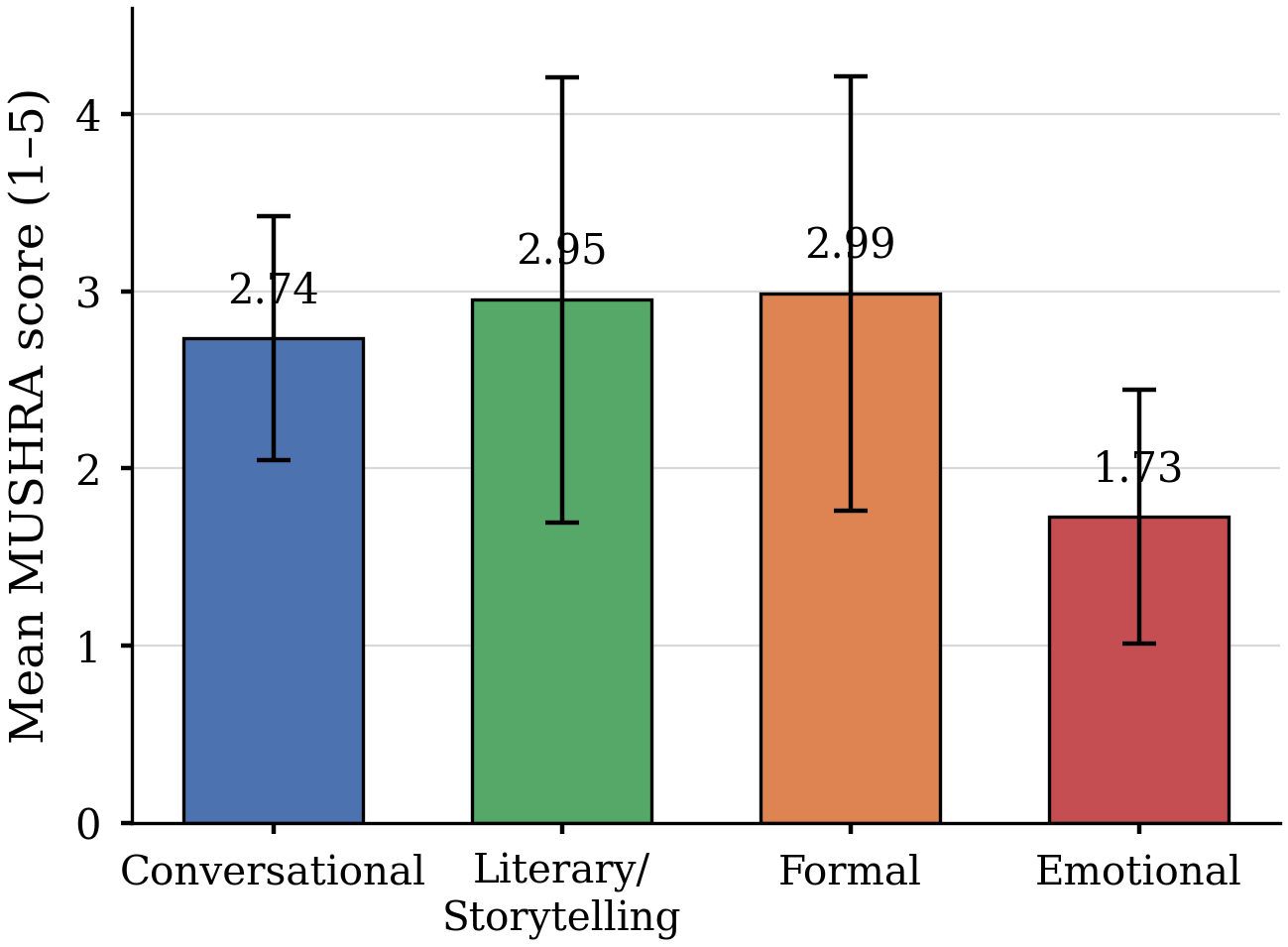}
    \caption{Domain-level mean MUSHRA-style scores.}
    \label{fig:mushra_domain}
    \end{figure}

    \subsubsection{System-Level MUSHRA Results}
    
    \begin{center}
    \captionof{table}{Per-system mean MUSHRA-style scores (1--5 scale) by domain.
    Conv.\ = Conversational, Lit.\ = Literary/Storytelling, Form.\ = Formal,
    Emot.\ = Emotional. Bold = best system per domain (column).}
    \label{tbl:mushra_system}
    
    \footnotesize
    \setlength{\tabcolsep}{7.7pt}
    \begin{tabular}{@{}lccccc@{}}
    \toprule
    \textbf{System} & \textbf{Conv.} & \textbf{Lit.} & \textbf{Form.} &
    \textbf{Emot.} & \textbf{Overall} \\
    \midrule
    Edge TTS    & 3.093          & 3.361          & \textbf{3.373} & 1.963          & 2.947 \\
    Gemini TTS  & \textbf{3.148} & \textbf{3.602} & 3.244          & \textbf{2.167} & \textbf{3.040} \\
    Indic-Parler-TTS & 2.148     & 2.214          & 2.738          & 1.296          & 2.099 \\
    MMS-TTS     & 2.556          & 2.631          & 2.619          & 1.481          & 2.322 \\
    \bottomrule
    \end{tabular}
    
    \vspace{2.5mm}
    \footnotesize\textit{Note:} SDs (in parentheses, per domain-system cell) --
    Edge TTS: 0.442, 1.071, 1.210, 0.666; Gemini TTS: 0.487, 1.097, 1.416, 0.739;
    Indic-Parler-TTS: 0.650, 1.103, 1.001, 0.496; MMS-TTS: 0.598, 1.232, 1.090, 0.569
    (order: Conv., Lit., Form., Emot.).
    \end{center}

    Gemini TTS remains the top-ranked system overall (3.040), narrowly ahead of
    Edge TTS (2.947), with Indic-Parler-TTS (2.099) and MMS-TTS (2.322) trailing
    substantially. Gemini TTS leads on three of four domains (Conversational,
    Literary/Storytelling, Emotional); Edge TTS leads only on Formal (3.373 vs.
    3.244). The Emotional domain again produces the lowest scores for every
    system, with even the best performer (Gemini, 2.167) falling well below the
    scale midpoint.
    
    \subsection{ABX Discrimination Results}\label{sec:results-abx}

    Listeners correctly identified the reference-matching system in 90.7\% of
    trials overall (98/108), well above the 50\% chance level in both domains.
    Accuracy was identical to one decimal place across Literary/Storytelling and
    Formal (49/54 each), indicating MMS-TTS and Gemini-TTS remain reliably
    distinguishable from the reference regardless of domain.
    
    \captionof{table}{ABX discrimination accuracy (\%), Edge-TTS vs. Gemini-TTS.
    Chance level is 50\%. $n=54$ trials per domain (9 listeners $\times$ 6
    samples).}\label{tbl:abx}
    
    \small
    \renewcommand{\arraystretch}{1.15}
    
    \begin{tabular*}{\columnwidth}{@{\extracolsep{\fill}}lcc@{}}
    \toprule
    \textbf{Domain} & \textbf{Accuracy (\%)} & \textbf{$n$} \\
    \midrule
    Literary/Storytelling & 90.7 & 54 \\
    Formal                & 90.7 & 54 \\
    \midrule
    \textbf{Overall}      & \textbf{90.7} & 108 \\
    \bottomrule
    \end{tabular*}

    \subsection{Speaker Similarity via Resemblyzer}\label{sec:results-speaker}

    \hspace*{1.2em} Speaker similarity results are analysed at the domain, system, and cross-domain levels. Domain-level analyses characterise overall voice identity preservation across speech domains, system-level analyses compare the performance of individual TTS systems, and cross-domain analyses assess the stability of system rankings across different domains.

    \subsubsection{Domain-Level Speaker Similarity}
    
    \hspace*{1.2em} Table~\ref{tbl:sim_domain} reports mean cosine similarity between reference and synthetic speaker embeddings aggregated across all four systems per
    domain. A consistent domain-level gradient is observed, with Literary/
    Storytelling achieving the highest mean similarity (0.6903) and the
    Emotional domain the lowest (0.5437), representing a difference of 0.1466
    cosine similarity units across the domain range.
    
    \begin{center}
    \captionof{table}{Domain-level mean Resemblyzer cosine similarity aggregated across
    all four TTS systems ($n = 240$ pairs per domain).}
    \label{tbl:sim_domain}
    
    \small
    \begin{tabular*}{\columnwidth}{@{\extracolsep{\fill}}p{3.0cm}p{2cm}p{1.0cm}@{}}
    \toprule
    \textbf{Domain} & \textbf{Mean} & \textbf{SD} \\
    \midrule
    Literary/Storytelling & 0.6903 & 0.0793 \\
    Conversational        & 0.6592 & 0.0739 \\
    Formal                & 0.5758 & 0.0765 \\
    Emotional             & 0.5437 & 0.0667 \\
    \bottomrule
    \end{tabular*}
    \end{center}

    \hspace*{1.2em} The domain ranking on speaker similarity differs from the domain rankings
    on MCD and F0 RMSE in one substantive respect: Literary/Storytelling ranks
    first on speaker similarity (0.6903) but second on MCD (7.87\,dB) and
    jointly first on F0 RMSE (521.24\,cents). Conversational ranks second on
    speaker similarity (0.6592) but first on MCD. This dissociation indicates
    that speaker identity preservation and spectral envelope fidelity are
    partially independent dimensions of synthesis quality and that a system
    may preserve the perceptual characteristics of a speaker's voice while
    still diverging acoustically from the reference signal at the frame level.
    The Formal domain ranks third on speaker similarity (0.5758) despite
    containing what might be expected to be the most predictable prosodic
    content, a pattern attributable to within-domain heterogeneity between
    its two source subsets, as elaborated in
    Section~\ref{sec:results-speaker-system}. The Emotional domain ranks
    last on speaker similarity (0.5437), consistent with its ranking on both
    MCD and F0 RMSE, confirming that emotional speech presents the most
    challenging conditions for synthesis quality preservation across all
    three objective dimensions.
    
    \subsubsection{System-Level Speaker Similarity}\label{sec:results-speaker-system}
    
    \hspace*{1.2em} Table~\ref{tbl:sim_system} reports cosine similarity per system per domain. All reference and synthetic audio is male-speaker. MMS-TTS produces a fixed single-speaker (male) output; because its voice identity is fixed and differs from each domain's reference speaker, its similarity scores primarily reflect speaker-identity mismatch rather than synthesis quality, and should be interpreted accordingly.
    
    \begin{table*}[t]
    \caption{Per-system Resemblyzer cosine similarity across four speech
    domains ($n = 60$ pairs per system-domain cell). Higher values indicate
    better speaker identity preservation. Bold denotes the best-performing
    system within each domain. All reference and synthetic audio is
    male-speaker; MMS-TTS uses a fixed single-speaker voice, so its scores
    reflect speaker-identity mismatch rather than gender
    mismatch.}\label{tbl:sim_system}
    \small
    \renewcommand{\arraystretch}{1.15}
    
    \begin{tabular*}{\textwidth}{@{\extracolsep{\fill}}lcccc@{}}
    \toprule
    \textbf{Domain} & \textbf{Edge TTS} & \textbf{Gemini TTS} &
    \textbf{Indic-Parler-TTS} & \textbf{MMS-TTS} \\
    \midrule
    Conversational        & 0.6916          & 0.6303          & \textbf{0.7293} & 0.5857 \\
    Literary/Storytelling & 0.7393          & 0.6676          & \textbf{0.7542} & 0.6000 \\
    Formal                & 0.5398          & 0.5647          & 0.5172          & \textbf{0.6814} \\
    Emotional             & 0.5610          & 0.5196          & 0.5241          & \textbf{0.5700} \\
    \midrule
    \textbf{Overall Mean} & 0.6329          & 0.5956          & \textbf{0.6312} & 0.6093 \\
    \bottomrule
    \end{tabular*}
    \end{table*}

    Table~\ref{tbl:sim_emotion} reports the cross-system emotion-level breakdown.

    \begin{table*}[t]
    \captionof{table}{Cross-system mean Resemblyzer cosine similarity by emotion
    within the Emotional domain ($n = 10$ pairs per system per emotion).
    Higher values indicate better speaker identity preservation. Emotions
    are listed in descending order of cross-system mean similarity.}
    \label{tbl:sim_emotion}
    
    \small
    \renewcommand{\arraystretch}{1.15}
    
    \begin{tabular*}{\textwidth}{@{\extracolsep{\fill}}lccccc@{}}
    
    \toprule
    \textbf{Emotion} & \textbf{Edge} & \textbf{Gemini} &
    \textbf{Indic} & \textbf{MMS} & \textbf{Mean} \\
    \midrule
    Sad       & 0.5828 & 0.5412 & 0.5375 & 0.5855 & \textbf{0.5618} \\
    Fear      & 0.5756 & 0.5343 & 0.5204 & 0.5759 & 0.5516 \\
    Anger     & 0.5662 & 0.5059 & 0.5618 & 0.5580 & 0.5480 \\
    Disgust   & 0.5383 & 0.5339 & 0.5155 & 0.5701 & 0.5395 \\
    Happy     & 0.6106 & 0.5103 & 0.4949 & 0.5243 & 0.5350 \\
    Boredom   & 0.5169 & 0.4926 & 0.5105 & 0.5931 & 0.5283 \\
    \bottomrule
    \end{tabular*}
    \end{table*}
    
    Boredom yields the lowest cross-system mean similarity (0.5283), a
    counterintuitive finding given that boredom is a low-arousal state
    with reduced prosodic variation. A plausible explanation is that
    boredom in Urdu speech is characterised by atypical pitch declination
    patterns and reduced loudness that deviate from neutral read-speech
    baselines in ways distinct from high-arousal emotions, making speaker
    embedding extraction less reliable even when the acoustic signal is
    relatively flat. Edge TTS is the only system that achieves its highest
    emotion-level similarity on Happy (0.6106), while all other systems
    rank Happy fourth or lower, suggesting that Edge TTS's neural vocoder
    better preserves speaker identity under conditions of increased
    breathiness and higher pitch that characterise happy speech in Urdu.
    
    \subsubsection{Cross-Domain Speaker Similarity Pattern}

    \hspace*{1.2em} Indic-Parler-TTS leads on Conversational and Literary but
    ranks last on Formal. MMS-TTS leads on Formal and Emotional but ranks
    last on Conversational and Literary. This cross-domain rank instability
    on speaker similarity mirrors the rank reversal observed on MCD in the
    Emotional domain (Section~\ref{sec:results-mcd}) and provides
    convergent evidence that system selection based on single-domain
    evaluation would systematically misrepresent deployment performance
    across communicative contexts. The practical implication is that no
    currently available Urdu TTS system provides uniformly high speaker
    identity preservation across the full range of speech domains evaluated
    in this study.
    
    \subsection{Mel-Cepstral Distortion}\label{sec:results-mcd}
    \hspace*{1.2em} MCD results are analysed at both the domain and system levels. Domain-level analyses characterise overall spectral fidelity across speech domains, while system-level analyses compare the spectral reconstruction performance of individual TTS systems within each domain.

    \subsubsection{Domain-Level MCD Results}
    
    \hspace*{1.2em} Across all 960 pairs the global mean MCD was 8.98\,dB
    ($\sigma = 2.77$\,dB). Table~\ref{tbl:mcd_domain} reports domain-level
    aggregates.
    
    \begin{center}
    \captionof{table}{Domain-level mean MCD (dB) aggregated across all four TTS systems
    ($n = 240$ pairs per domain). Lower values indicate better spectral
    envelope fidelity.}\label{tbl:mcd_domain}
    \small
    \begin{tabular*}{\columnwidth}{@{\extracolsep{\fill}}lcc@{}}
    \toprule
    \textbf{Domain} & \textbf{Mean MCD (dB)} & \textbf{SD} \\
    \midrule
    Conversational        & 6.90 & 1.76 \\
    Literary/Storytelling & 7.87 & 1.56 \\
    Formal                & 9.12 & 1.79 \\
    Emotional             & 12.03 & 2.69 \\
    \bottomrule
    \end{tabular*}
    \end{center}
    
    \hspace*{1.2em} The Conversational domain achieves the lowest mean MCD (6.90\,dB), while the Emotional domain produces the highest (12.03\,dB), representing a 74.3\% increase in spectral distortion. The Literary/Storytelling domain occupies an intermediate position (7.87\,dB), ranking second despite containing narratively complex, stylistically marked speech, a finding elaborated in Section~\ref{sec:discussion-domain}. The Formal domain produces higher mean MCD (9.12\,dB) than Literary/Storytelling, attributable in part to within-domain heterogeneity arising from its two source subsets (FLEURS and UrduSpeech), as reflected in its elevated standard deviation (1.79\,dB) relative to Literary/Storytelling (1.56\,dB).
    
    \subsubsection{System-Level MCD Results}
    
    \hspace*{1.2em} Table~\ref{tbl:mcd_system} reports MCD per system per domain.

    \begin{table*}[t]
    \caption{Per-system mean MCD (dB) across four speech domains ($n = 60$
    pairs per system-domain cell). Lower values indicate better spectral
    envelope fidelity. Bold denotes the best-performing system within each
    domain.}\label{tbl:mcd_system}
    
    \small
    \renewcommand{\arraystretch}{1.15}
    
    \begin{tabular*}{\textwidth}{@{\extracolsep{\fill}}lcccc@{}}
    \toprule
    \textbf{Domain} & \textbf{Edge TTS} & \textbf{Gemini TTS} &
    \textbf{Indic-Parler-TTS} & \textbf{MMS-TTS} \\
    \midrule
    Conversational        & \textbf{5.62} & 7.30 & 6.44 & 8.25 \\
    Literary/Storytelling & \textbf{6.32} & 8.67 & 7.09 & 9.40 \\
    Formal (FLEURS)       & \textbf{7.98} & 11.15 & 10.37 & 10.37 \\
    Formal (UrduSpeech)    & \textbf{7.47} & 8.24 & 7.51 & 9.90 \\
    Emotional             & 12.17          & 13.58 & \textbf{10.62} & 11.75 \\
    \midrule
    \textbf{Overall Mean} & \textbf{7.91} & 9.79 & 8.41 & 9.93 \\
    \bottomrule
    \end{tabular*}
    \end{table*}

    Edge TTS achieves the lowest MCD in four of five domain-subset conditions
    (Conversational, Literary/Storytelling, and both Formal subsets),
    establishing it as the most spectrally accurate system overall, with a
    mean MCD of 7.91\,dB. MMS-TTS produces the highest overall mean MCD
    (9.93\,dB), a difference of 2.02\,dB relative to Edge TTS, and the highest
    MCD in the Conversational, Literary/Storytelling, and Formal (Humairawan)
    conditions. Within the Emotional domain, system rankings shift notably:
    Indic-Parler-TTS achieves the lowest MCD (10.62\,dB) despite ranking
    second overall, while Gemini TTS, which performs competitively in other
    domains, produces the highest Emotional MCD (13.58\,dB). This rank
    reversal on the most challenging domain is a substantive finding examined
    in Section~\ref{sec:discussion-system}.
    
    \subsection{Fundamental Frequency RMSE}\label{sec:results-f0}

    \hspace*{1.2em} F0 RMSE results are presented at the domain level before being disaggregated by system to examine the pitch contour accuracy of individual TTS models within each speech domain.

    \subsubsection{Domain-Level F0 RMSE Results}
    
    \hspace*{1.2em} The global mean F0 RMSE across all 960 pairs was 619.34\,cents
    ($\sigma = 229.44$\,cents), with a coefficient of variation of 0.371,
    higher than the MCD coefficient of variation of 0.309.
    This indicates that pitch accuracy is more variable across the
    system-domain space than spectral envelope fidelity. Table~\ref{tbl:f0_domain}
    reports domain-level aggregates.
    
    \begin{center}
    \captionof{table}{Domain-level mean F0 RMSE (cents) aggregated across all four TTS
    systems ($n = 240$ pairs per domain). Lower values indicate better pitch
    contour accuracy.}\label{tbl:f0_domain}
    
    \small
    \setlength{\tabcolsep}{7pt}
    \renewcommand{\arraystretch}{1.2}
    
    \begin{tabular*}{\columnwidth}{@{\extracolsep{\fill}}lcc@{}}
    \toprule
    \textbf{Domain} & \textbf{Mean F0 RMSE (cents)} & \textbf{SD} \\
    \midrule
    Literary/Storytelling & 521.24 & 89.63  \\
    Formal                & 520.98 & 95.67  \\
    Conversational        & 546.02 & 114.25 \\
    Emotional             & 888.54 & 288.75 \\
    \bottomrule
    \end{tabular*}
    \end{center}
    
    \hspace*{1.2em} The F0 RMSE domain ranking differs from the MCD ranking in one notable
    respect: Literary/Storytelling and Formal are ranked first and second
    respectively on F0 RMSE, whereas Conversational ranked first on MCD.
    This dissociation indicates that Conversational speech, while spectrally
    close to the reference, exhibits greater pitch trajectory deviation than
    Literary or Formal speech. A plausible explanation is that conversational
    Urdu employs more varied intonational patterns than read or narrated speech,
    producing larger F0 deviations even when the spectral character of the
    voice is well preserved. The Emotional domain again stands apart, with a
    mean F0 RMSE of 888.54\,cents representing a 70.4\% increase over the
    next highest domain (Conversational at 546.02\,cents). The standard
    deviation of 288.75\,cents on the Emotional domain is more than three
    times that of Formal (95.67\,cents), reflecting the high inter-utterance
    variability inherent in emotionally marked speech.
    
    \subsubsection{System-Level F0 RMSE Results}
    
    \hspace*{1.2em} Table~\ref{tbl:f0_system} reports F0 RMSE per system per domain.

    \begin{table*}[t]
    \caption{Per-system mean F0 RMSE (cents) across four speech domains
    ($n = 60$ pairs per system-domain cell). Lower values indicate better
    pitch contour accuracy. Bold denotes the best-performing system within
    each domain.}\label{tbl:f0_system}
    
    \small
    \renewcommand{\arraystretch}{1.15}
    
    \begin{tabular*}{\textwidth}{@{\extracolsep{\fill}}lcccc@{}}
    \toprule
    \textbf{Domain} & \textbf{Edge TTS} & \textbf{Gemini TTS} &
    \textbf{Indic-Parler-TTS} & \textbf{MMS-TTS} \\
    \midrule
    Conversational        & \textbf{500.95} & 555.40 & 592.89 & 535.77 \\
    Literary/Storytelling & \textbf{505.53} & 537.21 & 550.12 & 492.08 \\
    Formal (FLEURS)       & 570.77          & 620.41 & \textbf{470.66} & 470.66 \\
    Formal (UrduSpeech) & \textbf{482.63} & 490.04 & 541.34 & 521.31 \\
    Emotional             & \textbf{765.14} & 1027.44 & 915.53 & 846.05 \\
    \midrule
    \textbf{Overall Mean} & \textbf{565.00} & 646.10 & 614.11 & 573.17 \\
    \bottomrule
    \end{tabular*}
    \end{table*}
    
    Edge TTS achieves the lowest F0 RMSE in four of five domain-subset
    conditions, with an overall mean of 565.00\,cents. MMS-TTS ranks second
    overall (573.17\,cents), closely trailing Edge TTS despite ranking last
    on MCD, indicating a dissociation between spectral and prosodic accuracy
    for this system. Gemini TTS produces the highest F0 RMSE in the Emotional
    domain (1027.44\,cents), exceeding the chance-level pitch deviation
    expected from a system generating unconditioned prosody, and ranking
    last overall on this metric (646.10\,cents). Within the Formal FLEURS
    subset, Indic-Parler-TTS and MMS-TTS produce identical F0 RMSE values
    (470.66\,cents), suggesting that both systems converge on a similar
    pitch strategy for read-speech prompts from this source, likely reflecting
    shared training on formal read-speech data.

    \subsection{Cross-Metric Synthesis}\label{sec:results-synthesis}
    
    \hspace*{1.2em} The preceding subsections reported each evaluation paradigm independently. Considered jointly, however, the evaluation framework captures three complementary dimensions of TTS performance: a \textit{holistic perceptual} quality (MUSHRA, ABX), a \textit{spectral-acoustic fidelity} (MCD), and a \textit{prosodic} and \textit{speaker identity preservation} (F0 RMSE and Resemblyzer similarity, respectively). The central question is not which system wins on aggregate, but where these axes agree, where they diverge, and what the divergence reveals about what each method actually measures.
    
    \subsubsection{Convergence at the Domain Level}
    
    \hspace*{1.2em} All five paradigms agree on one point: the Emotional domain is the most difficult condition in this corpus. MUSHRA scores are lowest for every system on Emotional (Table~\ref{tbl:mushra_system}), MCD is highest for
    every system on this domain (12.03\,dB at the domain level, against a
    corpus mean of 8.98\,dB), F0 RMSE is more than 1.6 times higher than
    the next-worst domain, and Resemblyzer similarity is at its global minimum
    (0.5437). This five-way convergence is notable precisely because the
    paradigms differ so substantially in what they operationalize: MUSHRA
    reflects integrated human preference, MCD reflects frame-level spectral
    match, F0 RMSE reflects pitch trajectory accuracy, and Resemblyzer
    reflects speaker-embedding distance. Agreement across all four
    \emph{independent} measurement constructs constitutes considerably
    stronger evidence for the Emotional domain's difficulty than any single
    metric could provide in isolation. This pattern implies that affective
    prosody is not merely under-rated by listeners; it is genuinely harder
    to reproduce at the acoustic level, and the difficulty is multidimensional
    rather than confined to pitch, timbre, or spectral shape alone.
    
    \subsubsection{Divergence at the System Level}
    
    \hspace*{1.2em} The system-level picture is considerably less consistent, and the disagreements are informative. Gemini TTS achieves the highest overall MUSHRA score (3.040) yet the highest (worst) overall F0 RMSE (646.10\,cents) and the lowest overall Resemblyzer similarity (0.5956) among the four systems; it additionally produces the highest Emotional-domain MCD
    (13.58\,dB) and the highest Emotional-domain F0 RMSE (1027.44\,cents) of
    any system in this study. In other words, the system that listeners
    preferred most was, on every objective acoustic measure examined, the
    system whose output deviated most from the reference signal. This is not
    a small or marginal discrepancy; it is the single largest subjective–objective inversion in the dataset.
    
    \begin{sloppypar}
    \hspace*{1.2em} A plausible explanation, consistent with established findings in expressive speech synthesis, is that MUSHRA-style ratings capture
    \textit{perceived naturalness and well-formedness}, not
    \textit{reference fidelity}. Gemini TTS likely generates fluent,
    confidently articulated prosodic contours that sound natural in isolation
    even when they do not track the idiosyncratic pitch excursions or
    spectral irregularities of the specific reference recording -- irregularities
    that, in the Emotional domain particularly, may themselves be partially
    attributable to expressive variability rather than to a canonical ``correct''
    target. Listeners evaluating a synthesized clip against a reference and an
    anchor in a MUSHRA paradigm are not penalized for prosodic divergence per se;
    they are rewarded for sounding convincing. MCD and F0 RMSE, by contrast,
    penalize any departure from the reference trajectory, regardless of whether
    that departure sounds more or less natural to a listener. The divergence
    between the perceptual and signal-fidelity metrics therefore indicates that
    reference-matching and perceived quality are partially orthogonal
    constructs for emotionally expressive speech: a system can sound
    convincingly emotional while diverging substantially, in cents and decibels,
    from the specific acoustic realization a human speaker happened to produce.
    \end{sloppypar}
    
    \hspace*{1.2em} A complementary disagreement appears for Indic-Parler-TTS, which ranks lowest or near-lowest on MUSHRA in three of four domains
    (Table~\ref{tbl:mushra_system}) yet achieves the highest Resemblyzer
    similarity overall (0.6312) and the lowest Emotional-domain MCD
    (10.62\,dB). Here the inversion runs in the opposite direction: a system
    that is objectively close to the reference, both spectrally and in speaker
    identity, is nonetheless rated poorly by listeners. This pattern is
    consistent with Indic-Parler-TTS's description-conditioned architecture
    producing output that matches the coarse spectral envelope and speaker
    embedding of the reference while exhibiting local artifacts, monotone
    delivery, or unnatural micro-prosody that frame-level and embedding-level
    metrics are not designed to detect. Embedding-based similarity and
    cepstral distance operate at a granularity -- global spectral shape,
    fixed-dimensional speaker representation -- that does not capture the kind
    of moment-to-moment naturalness defects that drive human dispreference.
    Conversely, MUSHRA captures exactly these defects but provides no
    diagnostic insight into \textit{why} a system sounds unnatural, since it
    returns a single integrated quality judgment.
    
    \hspace*{1.1em} The ABX results add a further dimension that neither MUSHRA nor the acoustic metrics provide. Even the systems involved in the ABX comparison (Edge-TTS and Gemini TTS) remain discriminable from the reference in 90.7\% of trials in both domains tested, despite Gemini TTS's relatively strong MUSHRA performance. This confirms that perceived \textit{quality} and
    perceived \textit{authenticity} are separable constructs: a synthesis can
    be rated as good or even excellent on a quality scale while still being
    reliably identifiable as non-human in a forced-choice discrimination task.
    This distinction is consequential for deployment contexts in which the
    goal is not merely high quality but indistinguishability from human speech
    (e.g., voice cloning, dubbing); MUSHRA alone would systematically overstate
    progress toward that goal.
    
    \subsubsection{Complementary Coverage Across Paradigms}
    
    \begin{sloppypar}
    \hspace*{1.2em} Taken together, no single paradigm in this study subsumes the others. MUSHRA captures integrated human preference but conflates multiple underlying causes of (dis)satisfaction into one score and cannot localize whether a deficit originates in pitch, timbre, segmental quality, or
    speaker identity. ABX isolates discriminability from naturalness, a
    distinction MUSHRA cannot make on its own. MCD isolates segmental spectral
    accuracy but is insensitive to global naturalness, as the Gemini case
    demonstrates. F0 RMSE isolates prosodic contour accuracy and is the most
    volatile metric across domains (coefficient of variation 0.371 versus
    0.309 for MCD), consistent with pitch being the acoustic dimension most
    sensitive to expressive content and least constrained by training-data
    priors. Resemblyzer isolates speaker identity preservation independent of
    content and prosody, but, as the Indic-Parler-TTS and MMS-TTS
    cases illustrate, can be confounded by architectural properties (a fixed
    single-speaker output, in the case of MMS-TTS) unrelated to synthesis
    quality. 
    \end{sloppypar}
    
    \subsubsection{Domain-Specific Cross-Metric Effects}
    
    \hspace*{1.2em} The Formal domain illustrates a further complication that aggregate domain-level statistics conceal. Its two constituent subsets, FLEURS and UrduSpeech, produce divergent system orderings on both MCD and F0 RMSE (Tables~\ref{tbl:mcd_system} and~\ref{tbl:f0_system}): Indic-Parler-TTS and MMS-TTS tie for the lowest F0 RMSE on FLEURS but Indic-Parler-TTS ranks
    worst on speaker similarity for the domain overall. This within-domain
    heterogeneity, also visible in Edge TTS's split between UrduSpeech and
    FLEURS subsets in Section~\ref{sec:results-speaker} (a 0.2324 similarity
    gap between subsets), indicates that "Formal" is not acoustically
    homogeneous in this corpus, and that conclusions drawn from its
    domain-level aggregate risk masking subset-specific effects that are
    themselves as large as the cross-domain effects the study set out to
    measure.
    
    \subsubsection{Integrated Interpretation}
    
    \hspace*{1.2em} Collectively, these findings suggest that no evaluation paradigm employed here, taken alone, would have produced a complete or even directionally reliable account of system quality. MUSHRA alone would have crowned Gemini TTS the unambiguous winner; the objective metrics alone would have crowned Edge TTS instead; Resemblyzer alone would have favored
    Indic-Parler-TTS for non-Formal domains. The disagreement is not noise --
    it recurs consistently across the Emotional domain in particular, where
    every paradigm agrees the task is hard but no paradigm agrees on who
    handles the difficulty best. This indicates that "best TTS system" is not
    a domain- or metric-invariant property in this evaluation, and that any
    single-number leaderboard claim for Urdu TTS would be an artifact of
    which evaluation paradigm, and which domain, was chosen to report it.
    
    \section{Discussion}\label{sec:discussion}

    \hspace*{1.2em} This section interprets the evaluation results presented in Section~\ref{sec:results} and discusses their broader implications for the evaluation of modern text-to-speech (TTS) systems. The findings demonstrate that synthesis quality varies substantially across systems and speech domains, while different evaluation metrics often capture complementary aspects of performance rather than yielding consistent rankings. These observations highlight the importance of multi-metric evaluation and provide insights into the strengths and limitations of current TTS systems, with Urdu serving as a representative low-resource case study.

    \subsection{System-Level Findings}\label{sec:discussion-system}
    
    \hspace*{1.2em} The cross-metric synthesis indicates that system performance in this study is paradigm-dependent rather than uniform, and this dependency is itself the most informative system-level finding. Edge TTS is the only system that performs strongly across \textit{all} objective measures (lowest
    overall MCD and F0 RMSE, Tables~\ref{tbl:mcd_system}
    and~\ref{tbl:f0_system}) while remaining competitive, though not dominant,
    on MUSHRA. This profile is consistent with a synthesis strategy that
    tracks the reference acoustic envelope and pitch trajectory closely,
    which objective metrics reward directly, while producing output that
    listeners judge as good but not exceptional -- plausibly because
    conservative, closely-tracked prosody reads as competent rather than
    expressive.
    
    \hspace*{1.2em} Gemini TTS exhibits the inverse profile: weakest or near-weakest on every
    objective measure, yet strongest on MUSHRA (3.040 overall). This is the
    single largest subjective–objective inversion observed in the study. A
    plausible explanation is that Gemini TTS prioritizes fluent, confident
    delivery over strict reference tracking, a trade-off that pays off in
    human preference judgments but does not register as a fidelity gain on
    signal-level metrics, which penalize any departure from the reference
    trajectory regardless of whether that departure sounds more natural.
    
    \hspace*{1.2em} Indic-Parler-TTS occupies an unusual middle position: objectively competitive on speaker identity (highest overall Resemblyzer similarity, 0.6312) and, on the Emotional domain specifically, on spectral fidelity (lowest MCD among all systems, 10.62\,dB), yet consistently the
    lowest-rated or near-lowest-rated system on MUSHRA in three of four
    domains. This indicates that its description-conditioned architecture
    reproduces the coarse spectral envelope and speaker embedding of the
    reference while exhibiting local artifacts or unnatural micro-prosody
    that frame-level and embedding-level metrics are not designed to detect.
    MMS-TTS, meanwhile, is the most domain-conditional system: it ranks
    highest on speaker similarity in the Formal domain, consistent with
    training data dominated by formal read-speech, but last on Conversational
    and Literary similarity and consistently worst on MCD across all domains.
    
    \hspace*{1.2em} Considered together, these four profiles indicate that none of the systems evaluated has converged on a strategy that simultaneously
    satisfies signal-fidelity and listener-preference criteria. Each makes a
    different, identifiable trade-off, and that trade-off -- not a single
    quality ranking -- is the more accurate description of the current state
    of these systems for Urdu speech synthesis.
    
    \subsection{Domain-Level Variation}\label{sec:discussion-domain}
    
    \hspace*{1.2em} All five evaluation paradigms agree that the Emotional domain is the most difficult condition in this corpus: MUSHRA scores are lowest for every system, MCD and F0 RMSE are highest for every system, and Resemblyzer similarity is at its global minimum (0.5437). This convergence across methodologically independent constructs -- holistic preference,
    spectral distance, pitch-contour distance, and embedding distance --
    constitutes stronger evidence for the domain's difficulty than any single
    metric could provide, and indicates that affective prosody is not merely
    under-rated by listeners but is genuinely harder to reproduce at the
    acoustic level, across multiple independent acoustic dimensions
    simultaneously.
    
    \hspace*{1.2em} The reasons for this difficulty are plausibly compounding rather than singular. Emotional reference speech itself exhibits substantially higher acoustic variability than the other three domains, evident in its larger standard deviations across every metric (e.g., F0 RMSE SD of 288.75\,cents versus 95.67\,cents for Formal). A noisier, more variable reference target is harder for any synthesis system to track, and is also harder for human listeners to use as a stable comparison point -- a factor independently implicated in the comparatively reduced ABX discrimination rate observed for Edge TTS on this domain. The domain's difficulty therefore appears to arise jointly from the synthesis task and the evaluation task, rather than from synthesis quality alone.
    
    \hspace*{1.2em} The remaining three domains show a more nuanced pattern that aggregate domain-level statistics partially obscure. Conversational speech achieves the lowest MCD of any domain but does not achieve the lowest F0 RMSE, indicating that conversational Urdu is spectrally well-matched but
    prosodically more variable than read or narrated speech, plausibly
    reflecting the more varied intonational patterns characteristic of
    spontaneous dialogue. Literary/Storytelling, despite its narratively
    complex and stylistically marked content, ranks favorably on MCD and F0
    RMSE alike, and achieves the highest Resemblyzer similarity of any
    domain -- consistent with sustained, evenly-paced narrative speech
    providing more stable acoustic context for both synthesis and embedding
    extraction than shorter or more variable utterance types.
    
    \hspace*{1.2em} The Formal domain illustrates a further complication: its two source subsets, FLEURS and UrduSpeech, produce divergent system orderings on both MCD and F0 RMSE, and Edge TTS shows a 0.2324 similarity gap between its UrduSpeech and FLEURS subset performance. This within-domain heterogeneity indicates that "Formal" is not acoustically homogeneous in this corpus, and that domain-level aggregates for Formal risk masking subset-specific effects that are themselves as large as the cross-domain effects the study set out to measure. This pattern reinforces the broader point that domain labels in this evaluation are useful organizing categories but should not be treated as guarantees of internal acoustic homogeneity.
    
    \subsection{Metric Agreement and Disagreement Analysis}\label{sec:discussion-agreement}
    
    \hspace*{1.2em} The most consequential pattern in this study is the dissociation between what MUSHRA measures and what the objective acoustic metrics measure. Gemini TTS diverges from the reference signal more than any other system on MCD, F0 RMSE, and Resemblyzer similarity, yet is the most preferred system on MUSHRA. This divergence indicates that reference-matching and perceived quality are partially orthogonal constructs for expressive Urdu speech: a system can sound convincingly natural while diverging substantially, in cents and decibels, from the specific acoustic realization a human reference speaker happened to produce. The implication generalizes beyond this dataset -- a synthesis system optimized to minimize a reference-matching loss may be optimizing for a target that is only loosely coupled to the naturalness judgments that ultimately determine perceived quality.
    
    \hspace*{1.2em} A complementary disagreement runs in the opposite direction for Indic-Parler-TTS, which is objectively close to the reference, both spectrally and in speaker identity, yet rated poorly by listeners. Taken together with the Gemini TTS case, these two inversions show that neither subjective nor objective evaluation alone is sufficient: MUSHRA captures naturalness defects that frame-level and embedding-level metrics cannot detect, since cepstral distance and fixed-dimensional speaker embeddings operate at a granularity that does not register local artifacts or unnatural micro-prosody; conversely, the acoustic metrics capture systematic reference deviation that an integrated MUSHRA score cannot localize or explain. Each paradigm is sensitive to defects the other is structurally unable to detect.
    
    \hspace*{1.2em} The ABX results add a dimension that neither MUSHRA nor the acoustic metrics provide. Even Gemini TTS, the strongest system on MUSHRA, remains discriminable from the reference in 87.5--88.9\% of Emotional and
    Literary/Storytelling trials. This confirms that perceived
    \textit{quality} and perceived \textit{authenticity} are separable: a
    synthesis can be rated highly on a quality scale while remaining reliably
    identifiable as non-human in a forced-choice discrimination task. A
    quality-only evaluation would systematically overstate progress toward
    indistinguishability from human speech, a distinction that matters
    directly for applications such as voice cloning or dubbing where
    authenticity, not merely quality, is the deployment criterion.

    \subsection {Key Findings}
\hspace*{1.2em} The key findings of this study are as follows:
\begin{itemize}

\item \textbf{TTS system performance is inherently multi-dimensional.} No single system achieves consistently superior performance across perceptual quality, acoustic fidelity, and speaker similarity measures, demonstrating that different evaluation criteria capture complementary aspects of synthesis quality.
\item \textbf{Subjective naturalness and objective reference similarity are not always aligned.} The results show clear cases where systems preferred by listeners differ substantially from those achieving the closest acoustic match to reference speech, indicating that perceived quality extends beyond signal-level similarity.
\item \textbf{Emotional speech remains the most challenging synthesis domain.} Across subjective and objective evaluations, emotional speech consistently exhibits the lowest perceptual quality, highest acoustic distortion, largest pitch variation errors, and weakest speaker similarity, highlighting the difficulty of modeling expressive prosody.
\item \textbf{Modern TTS systems exhibit distinct optimization trade-offs.} Different systems demonstrate different strengths, with some favoring acoustic and speaker fidelity while others achieve higher listener preference through more natural and fluent speech generation.
\item \textbf{Speech quality and speech authenticity represent different evaluation objectives.} ABX discrimination results show that highly rated synthetic speech can still be reliably distinguished from natural speech, suggesting that current systems have improved naturalness without fully achieving human-level authenticity.
\item \textbf{Comprehensive TTS evaluation requires multiple complementary metrics.} Combining perceptual, discriminative, acoustic, and speaker-similarity analyses provides a more complete understanding of synthesis performance than any individual evaluation measure.
\end{itemize}
    
    
    \section{Conclusion}\label{sec:conclusion}
\hspace*{1.2em} This study addressed the need for systematic and comprehensive evaluation of modern text-to-speech systems across diverse communicative domains. To this end, it introduced a domain-stratified evaluation framework covering Formal, Conversational, Literary/Storytelling, and Emotional speech, and evaluated four state-of-the-art TTS systems using complementary subjective and objective measures, including MUSHRA-style listening tests, ABX discrimination, speaker similarity analysis with Resemblyzer, and acoustic evaluations based on MCD and F0~RMSE. The framework provides a structured approach for analyzing how synthesis quality varies across different speech styles and communicative settings.

\hspace*{1.2em} The results demonstrate that no single system consistently achieves the best performance across all evaluation dimensions and speech domains. While Gemini TTS obtains the highest overall MUSHRA-style score (3.040), indicating stronger perceived naturalness and overall quality, Edge TTS achieves the strongest objective acoustic performance, with the lowest mean MCD (7.91\,dB) and lowest mean F0~RMSE (565.00\,cents). Across all evaluated systems, Emotional speech emerges as the most challenging synthesis condition, exhibiting the lowest perceptual scores, highest acoustic distortion (MCD of 12.03\,dB and F0~RMSE of 888.54\,cents), and lowest speaker similarity (0.5437). The ABX results further indicate that current TTS systems remain distinguishable from natural speech, with listeners correctly identifying the reference-matching condition in 90.7\% of trials, highlighting remaining challenges in achieving human-level expressive synthesis.

\hspace*{1.2em} The findings reveal several broader insights into the evaluation of modern TTS systems. First, synthesis quality is inherently multi-dimensional, and system performance depends on the evaluation criterion considered. Second, subjective naturalness and objective reference similarity are not always aligned, demonstrating that perceptual quality cannot be fully characterized by signal-level fidelity alone. Third, current TTS systems exhibit distinct optimization trade-offs, with different models prioritizing naturalness, acoustic similarity, or speaker identity. Finally, the results show that perceived quality and authenticity are separate evaluation objectives, as highly rated synthetic speech can still be distinguishable from natural speech.

\hspace*{1.2em} Overall, this study contributes: (i) a domain-stratified benchmark for evaluating TTS systems across diverse speech conditions; (ii) a multi-metric evaluation framework that integrates perceptual, acoustic, prosodic, and speaker-similarity analyses; and (iii) a publicly released evaluation pipeline, result tables, and reproducible resources that can support future research on speech synthesis evaluation. By demonstrating the importance of domain-aware and multi-dimensional assessment, this work provides a foundation for more comprehensive and context-sensitive evaluation of TTS systems, particularly in low-resource and underrepresented language settings.
        
    
    \section{Future Work}\label{sec:futurework}
    
\hspace*{1.2em} Future work can extend this benchmark in several directions. First, larger and more diverse listener studies using standardized online MUSHRA platforms would further improve the reliability and comparability of subjective evaluations. Second, the framework can be expanded to include additional speech scenarios and emerging TTS systems to investigate how synthesis quality varies across different communicative contexts and technological approaches. Finally, applying the proposed evaluation framework to other low-resource and underrepresented languages would help establish more inclusive and standardized practices for speech synthesis evaluation beyond high-resource settings.

    \section*{Data Set}
\hspace*{1.2em} The annotated evaluation data, including subjective listening test responses and benchmark results, is publicly available at: \url{https://github.com/Shifa402/Urdu-Synthetic-Speech-Evaluation}
The annotated dataset used in this study can be accessed with prior permission via this link: \url{https://drive.google.com/file/d/1MfeXJ-LWh1SXptU5gUbXckh27YLMURNW/view?usp=drive_link}
    \section*{Declaration of generative AI technologies in the writing process}
\hspace*{1.2em} During the preparation of this work the authors used Claude in order to improve the readability and language of the manuscript. After using this tool/service, the authors reviewed and edited the content as needed and take full responsibility for the content of the published article.

    
    \printcredits
    \nocite{*}
    \bibliographystyle{cas-model2-names}
    \bibliography{cas-refs}
    
    \end{document}